\documentclass{article}

\usepackage{microtype}
\usepackage{graphicx}
\usepackage{subfigure}
\usepackage{booktabs} 
\usepackage{longtable}

\usepackage{hyperref}

\usepackage[accepted]{icml2023}

\usepackage{amsmath}
\usepackage{amssymb}
\usepackage{mathtools}
\usepackage{amsthm}

\usepackage{times}
\usepackage{latexsym}
\usepackage{tabularx}
\usepackage{colortbl}
\usepackage{caption}
\usepackage{adjustbox}
\usepackage{multirow}
\usepackage{soul}
\usepackage{amsfonts}
\usepackage{bbding}
\usepackage{bbm}
\usepackage{xcolor}
\usepackage[normalem]{ulem}
\useunder{\uline}{\ul}{}

\usepackage[capitalize,noabbrev]{cleveref}

\theoremstyle{plain}

\theoremstyle{definition}

\theoremstyle{remark}

\usepackage[textsize=tiny]{todonotes}

\icmltitlerunning{Synthetic Prompting: Generating Chain-of-Thought Demonstrations for Large Language Models
}

\begin{document}

\twocolumn[
\icmltitle{Synthetic Prompting: Generating Chain-of-Thought Demonstrations for Large Language Models
}



\icmlsetsymbol{equal}{*}

\begin{icmlauthorlist}
\icmlauthor{Zhihong Shao}{tsinghua,intern}
\icmlauthor{Yeyun Gong}{msra}
\icmlauthor{Yelong Shen}{msr}
\icmlauthor{Minlie Huang}{tsinghua}
\icmlauthor{Nan Duan}{msra}
\icmlauthor{Weizhu Chen}{msr}
\end{icmlauthorlist}

\icmlaffiliation{tsinghua}{Tsinghua University}
\icmlaffiliation{intern}{This work was done during an internship in MSRA}
\icmlaffiliation{msra}{Microsoft Research Asia}
\icmlaffiliation{msr}{Microsoft}

\icmlcorrespondingauthor{Minlie Huang}{aihuang@tsinghua.edu.cn}

\icmlkeywords{Machine Learning, ICML}

\vskip 0.3in
]



\printAffiliationsAndNotice{}  

\begin{abstract}
Large language models can perform various reasoning tasks by using chain-of-thought prompting, which guides them to find answers through step-by-step demonstrations.
However, the quality of the prompts depends on the demonstrations given to the models, and creating many of them by hand is costly.
We introduce S\textsc{ynthetic prompting}, a method that leverages a few handcrafted examples to prompt the model to generate more examples by itself, and selects effective demonstrations to elicit better reasoning.
Our method alternates between a backward and forward process to generate new examples.
The backward process generates a question that match a sampled reasoning chain, so that the question is solvable and clear.
The forward process produces a more detailed reasoning chain for the question, improving the quality of the example.
We evaluate our method on numerical, symbolic, and algorithmic reasoning tasks, and show that it outperforms existing prompting techniques.
\end{abstract}

\section{Introduction}

Few-shot demonstrations, i.e., examples of inputs and outputs for a task, can enable Large Language Models (LLMs) to perform various tasks without fine-tuning \cite{DBLP:conf/nips/BrownMRSKDNSSAA20,DBLP:journals/corr/abs-2210-11416}. LLMs can further improve their performance by using chain-of-thought prompting, which provides intermediate reasoning steps for the task \cite{DBLP:journals/corr/abs-2201-11903,DBLP:journals/corr/abs-2205-11916}. However, the LLMs' few-shot performance depends heavily on the quality of the demonstrations, especially for reasoning tasks that need complex and diverse reasoning patterns. Manually creating a large and diverse set of examples for demonstration selection is costly and tedious, while relying on a limited set of demonstrations may hamper the LLMs' generalization and adaptation to different test inputs.

In this paper, we propose a novel method, S\textsc{ynthetic prompting}, that leverages the LLMs' own knowledge and generative power to augment a limited set of demonstrations with self-synthesized examples, and then uses the augmented set to elicit better reasoning in the LLMs. Specifically, given a few seed examples, each consisting of a question and a chain of reasoning steps, we prompt an LLM to generate more examples by alternating between two processes: (1) the backward process, where the LLM synthesizes a question based on a self-generated reasoning chain, which ensures that the question is answerable and well-defined; and (2) the forward process, where the LLM produces a reasoning chain for the synthesized question, which refines the reasoning chain to be more precise and consistent with the question. We repeat this process until we obtain enough synthetic examples. To select the most effective demonstrations from the augmented set, we propose a new selection scheme based on in-cluster complexity, which aims to maximize the diversity and informativeness of the demonstrations by clustering them and choosing the most complex one (the one with the longest reasoning chain) from each cluster. Finally, we prompt the LLM with the selected demonstrations to generate a reasoning chain for a test question and then use it to obtain the answer.

We evaluate our method on various reasoning tasks, including numerical reasoning, algorithmic reasoning, and symbolic reasoning. Following previous few-shot settings \cite{wang2022supernaturalinstructions,DBLP:journals/corr/abs-2210-09261},  we demonstrate that our method can significantly improve the LLMs' performance, achieving up to 15.6\% absolute gains over the state-of-the-art methods.

Our main contributions are:
\begin{itemize}
    \item We introduce S\textsc{ynthetic prompting}, a novel method that augments a limited set of demonstrations with self-synthesized examples by prompting an LLM, and leverages the augmented set to elicit better reasoning in the LLM.
    \item We propose an in-cluster complexity based scheme to select diverse and informative demonstrations from the augmented set for inference.
    \item We demonstrate the effectiveness of our method on three reasoning tasks, achieving significant improvements over previous methods. 
\end{itemize}

\section{Related Work}
\noindent
\textbf{In-context few-shot learning}
With large-scale unsupervised pre-training, LLMs \cite{DBLP:conf/nips/BrownMRSKDNSSAA20,DBLP:journals/corr/abs-2204-02311,DBLP:journals/corr/abs-2205-01068} can learn to perform tasks by mimicking in-context demonstrations \cite{DBLP:conf/naacl/ShinLAKKKCLPHS22}.
To improve robustness to prompts, instruction tuning \cite{DBLP:journals/corr/abs-2203-02155,DBLP:conf/iclr/WeiBZGYLDDL22,DBLP:conf/iclr/SanhWRBSACSRDBX22,DBLP:journals/corr/abs-2210-11416} has been proposed, which trains a language model on diverse tasks to generate desirable outputs that follow given instructions.
With improved controllability, in-context learning-based applications flourish, including text generation \cite{DBLP:journals/corr/abs-2212-10077,gao2022rarr}, dialogue generation \cite{DBLP:journals/corr/abs-2201-08239}, and resource construction \cite{DBLP:conf/naacl/WestBHHJBLWC22}.

\noindent
\textbf{Prompting techniques for reasoning}
Instead of directly generating an answer, chain-of-thought prompting \cite{DBLP:journals/corr/abs-2201-11903} prompts LLMs to arrive at an answer after a step-by-step reasoning process, which largely improves performance on numerous reasoning tasks.
Following work like least-to-most prompting \cite{DBLP:journals/corr/abs-2205-10625}, self-ask \cite{DBLP:journals/corr/abs-2210-03350}, and decomposed prompting \cite{DBLP:journals/corr/abs-2210-02406} also shares the spirit of question decomposition, i.e., decomposing a complex question into a series of tractable sub-questions.
All these methods produce natural language reasoning steps, which struggle with calculations and symbolic manipulations.
Techniques like P\textsc{a}L prompting \cite{DBLP:journals/corr/abs-2211-10435} and program-of-thought prompting \cite{DBLP:journals/corr/abs-2211-12588} propose to improve natural language reasoning with structured code, showing significant improvements on arithmetic, symbolic and algorithmic tasks.

Orthogonal to prompting workflows, there is also work that explores what make an effective demonstration.
Metrics include
(1) diversity, which selects complementary demonstrations so that models can fuse different reasoning \cite{DBLP:journals/corr/arxiv.2206.02336, DBLP:journals/corr/abs-2211-13892} or be less biased by one type of reasoning \cite{DBLP:journals/corr/abs-2210-03493};
(2) reasoning complexity, which selects demonstrations with the highest reasoning complexity, and has been found to work well on numerical reasoning empirically \cite{DBLP:journals/corr/abs-2210-00720};
(3) similarity with a test input, which retrieves structurally \cite{DBLP:journals/corr/abs-2209-15003} or semantically \cite{DBLP:conf/acl-deelio/LiuSZDCC22} similar demonstrations.
To ensure both diversity and informativeness of demonstrations, we propose a selection scheme based on in-cluster complexity to choose the most complex examples from example clusters.
All these selection schemes assume access to a set of examples (whether annotated or not).

\noindent
\textbf{Knowledge distillation from LLMs}
Some researches distilled knowledge from LLMs into symbolic knowledge, e.g., structured commonsense knowledge \cite{DBLP:conf/naacl/WestBHHJBLWC22} or task-specific examples \cite{DBLP:journals/corr/abs-2201-05955,DBLP:journals/corr/abs-2202-07922,DBLP:journals/corr/abs-2210-11610}.
These researches have at least one of the following characteristics:
(1) assuming access to gold inputs from training sets without needing to generate them;
(2) distilling knowledge based on collaboration between workers and AI;
(3) using distilled knowledge for training.
By contrast, we assume access to only a few gold examples, automatically synthesize more examples by prompting an LLM, and study whether synthesized examples can be leveraged to better elicit reasoning in the model itself, without further training.

\begin{figure*}[t!]
    \centering
    \includegraphics[width=0.99\textwidth]{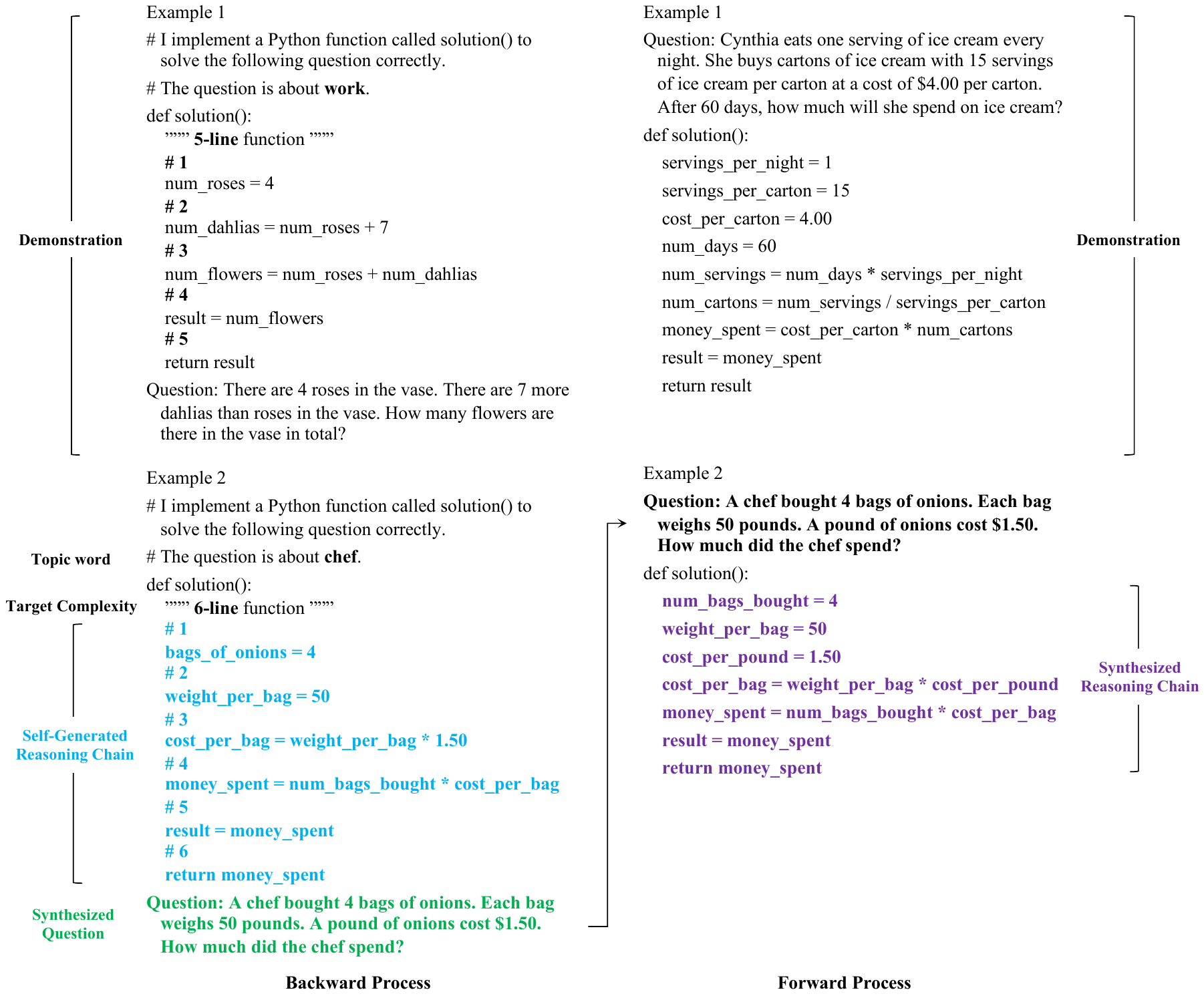}
    \caption{
    Example prompts and model completions in the backward process (left) and the forward process (right) of example synthesis.
    We show only one demonstration in each prompt for brevity.
    \texttt{Self-Generated Reasoning Chain} (in blue), \texttt{Synthesized Question} (in green), and \texttt{Synthesized Reasoning Chain} (in purple) are example completions.
    In the backward process, an LLM synthesizes a question conditioned on a topic word, a target reasoning complexity, and a generated reasoning chain.
    To better control the reasoning complexity, we number the reasoning steps, e.g., \texttt{\# 1} and \texttt{\# 2} on the left.
    In the forward process, the LLM synthesizes a more precise reasoning chain for the question produced in the backward process.
    The question produced in the backward process and the corresponding reasoning chain produced in the forward process constitute a synthetic example.
    }
    \label{fig:prompts_for_synthesis}
\end{figure*}

\section{Synthetic Prompting}
\subsection{Overview}
To perform reasoning tasks with LLMs, given a few examples each consisting of a question and a reasoning chain, it is common to directly concatenate them into a prompt for inference.
In this paper, we instead treat them as seed examples, and prompt an LLM to automatically synthesize more by repeating a backward-forward procedure; the backward process and the forward process produce a question and a corresponding reasoning chain, respectively.
During inference, the LLM is prompted with self-synthesized demonstrations to better elicit reasoning in the model itself.
Demonstrations are selected with a new scheme that ensures diversity and informativeness.

\subsection{Example Synthesis Phase}
Using seed demonstrations, we automatically synthesize more examples by repeating a backward-forward process.
Each synthetic example is a $\langle$ question, reasoning chain $\rangle$ pairs.
In our main experiments, we use P\textsc{a}L-style reasoning, i.e., reasoning chains are snippets of code, and answers are obtained by executing the code.

\subsubsection{Backward Process}
In the backward process, an LLM is prompted to first generate a reasoning chain and then a question.
The question, which is the output of the backward process, is synthesized conditioned on a given topic word, a target reasoning complexity, and the self-generated reasoning chain.
Figure \ref{fig:prompts_for_synthesis} (left) shows an example prompt for the backward process, which includes some demonstrations randomly sampled from the seed examples and the previously synthesized ones.
The number of demonstrations is equal to the number of seed examples.

\textbf{Topic word} We assume that each reasoning question is related to a specific topic, and that different topics may require different types of reasoning. For example, questions about \textit{tax} may involve arithmetic operations, while questions about \textit{speed} may involve unit conversions. To ensure diversity of the synthesized questions, we prompt the model to generate a question for a given topic word, which is randomly sampled from a set of words. The word set is created by prompting the model to list single-token noun words, following some random noun words from the seed examples. The instruction for generating the word set is \texttt{List 50 noun words. Each word should contain one token only. Do not repeat words already listed.}, followed by no more than 10 words from the seed examples.
We repeat this process until we have 1,000 different words, or reach 100 repetitions of prompting.

\textbf{Target complexity} We also want to control the complexity of the synthesized questions, as more complex examples may help the model learn better reasoning skills \cite{DBLP:journals/corr/abs-2210-00720}. We define the complexity of a question as the number of reasoning steps required to answer it, where a step is a line of code separated by a line break. For example, the complexity of \texttt{Example 1} in Figure \ref{fig:prompts_for_synthesis} (left) is 5, as it has 5 lines of code. The target complexity for generating a question is randomly sampled from a range that spans from the lowest complexity of the seed examples to the highest one plus $c$.

\textbf{Self-generated reasoning chain}
We prompt the model to generate a reasoning chain of the target complexity for the given topic, and then generate a question based on the reasoning chain. We find that this approach leads to more answerable and well-defined questions, compared to directly generating questions without a reasoning chain. To guide the model to follow the target complexity, we number each reasoning step in the demonstrations, e.g., \texttt{\#1} and \texttt{\#2} in Figure \ref{fig:prompts_for_synthesis} (left). We filter out the questions that are duplicated, repeat at least one 5-gram, or do not mention the given topic word.

\subsubsection{Forward Process}
The forward process aims to generate a reasoning chain for the question synthesized in the backward process. Figure \ref{fig:prompts_for_synthesis} (right) shows an example prompt for the forward process, which consists of the seed examples. Unlike chain-of-thought prompting, P\textsc{a}L prompting does not include the final answers in the prompt, as the answers can be obtained by executing the generated code, rather than extracted from the model output. We observe that the reasoning chain generated in the forward process is more relevant and precise than the one generated in the backward process, as it is directly conditioned on the question.

We also want to ensure that the model is confident about the answer produced by the reasoning chain. Following \citet{DBLP:journals/corr/abs-2210-11610}, we measure the confidence of an answer by the proportion of sampled reasoning chains that lead to the same answer. For a question $x$, we sample $m$ reasoning chains and obtain their answers $\{a_1, a_2, ..., a_m\}$. We then find the most consistent answer by majority voting: $\hat{a} = \arg max_{a_i} \sum_{k=1}^m \mathbbm{1}(a_i=a_k)$. If more than $m/2$ reasoning chains lead to $\hat{a}$, we associate the shortest one with the synthesized question; otherwise, we discard the question, as the model fails to produce confident reasoning chains for it. Note that majority voting is only used for synthesizing examples, not for inference (Section \ref{sec:infer}). This is different from \citet{DBLP:journals/corr/abs-2203-11171}, who use majority voting for inference.

\subsection{Inference Phase}
\label{sec:infer}
During inference, we select a subset of synthesized examples as demonstrations for the model. According to \citet{DBLP:journals/corr/abs-2210-00720}, selecting demonstrations based on complexity can improve the performance of the model on reasoning tasks, compared to selecting them based on similarity. Moreover, selecting demonstrations based on similarity may introduce biases \cite{DBLP:journals/corr/abs-2210-03493,DBLP:journals/corr/abs-2212-09865} from the demonstrations, especially if they are incorrect. Furthermore, selecting demonstrations that are complementary to each other may help the model fuse knowledge from different types of reasoning \cite{DBLP:journals/corr/abs-2211-13892,DBLP:journals/corr/abs-2210-03493}.

Therefore, we propose an in-cluster complexity based scheme to select demonstrations that are both complex and complementary. Specifically, we cluster the synthesized examples in a semantic embedding space, using Sentence-BERT \cite{reimers-gurevych-2019-sentence} as the encoder. The number of clusters is equal to the number of demonstrations used for inference. We then choose the most complex example from each cluster as the demonstration. The inference process is the same as previous work like P\textsc{a}L prompting, where the model completes a given prompt. The only difference is that the demonstrations in our prompts are synthesized from the seed examples, rather than fixed to them.

\begingroup
\setlength{\tabcolsep}{3pt} 
\renewcommand{\arraystretch}{1} 
\begin{table*}[htb]
    \centering
    \adjustbox{max width=.99\textwidth}{
    \begin{tabular}{p{0.2\textwidth}p{0.79\textwidth}}
        \hline
        Datasets & Example \\
        \hline
        GSM8K & Patrick has three glue sticks that are partially used. One has 1/6 left, the second has 2/3 left and the third one has 1/2 left. If a glue stick is 12 millimeters long originally, what is the total length of the glue sticks that are not used? \\
        Colored Objects & On the nightstand, you see several items arranged in a row: a blue crayon, a red notebook, a teal bracelet, a magenta sheet of paper, a silver dog leash, and a black booklet. What is the color of the item furthest from the dog leash? \\
        Repeat Copy & Repeat election to the council three times, but after every other word say cool \\
	\hline
    \end{tabular}
    }
    \caption{
    Examples from three datasets.
    Questions in the other numerical reasoning datasets resemble those in GSM8K.
    }
    \label{tab:dataset_samples}
\end{table*}
\endgroup

\section{Experiments}
\subsection{Datasets}
We experimented on seven datasets of different reasoning tasks.
Examples are presented in Table \ref{tab:dataset_samples}.

\textbf{Numerical reasoning}
(1) GSM8K \cite{DBLP:journals/corr/abs-2110-14168} is a dataset of 1,319 diverse grade school math word problems, curated to evaluate multi-step mathematical reasoning abilities of LLMs.
(2) GSM-Hard is a harder version of GSM8K, created by \citet{DBLP:journals/corr/abs-2211-10435} via replacing numbers in the questions with larger ones, intended to evaluate whether LLMs can generalize to large numbers.
(3) SVAMP \cite{DBLP:conf/naacl/PatelBG21} is a math word problem dataset with 1,000 questions for robustness evaluation.
(4) ASDiv \cite{DBLP:conf/acl/MiaoLS20} consists of 2,000 diverse math word problems.
(5) SingleOp \cite{DBLP:conf/naacl/Koncel-Kedziorski16} consists of 562 math word problems.

\textbf{Symbolic reasoning}
The Colored Objects task from Big-Bench Hard \cite{DBLP:journals/corr/abs-2210-09261}, with 2,000 questions about position and color attributes of given objects.

\textbf{Algorithmic reasoning}
The Repeat Copy task also comes from Big-Bench Hard, consisting of 32 test examples.
A model should generate a sequence of words that meets requirements in a given instruction.

\subsection{Evaluation Settings}
Both \citet{DBLP:journals/corr/abs-2210-09261} and \citet{wang2022supernaturalinstructions} evaluated LLMs on benchmarks with numerous tasks under few-shot settings which have access to no more than 4 gold examples.
Following these settings, we assumed access to 2 or 4 random examples from each dataset by default.
For numerical reasoning tasks, we also experimented with the 8 examples that were manually crafted by \citet{DBLP:journals/corr/abs-2201-11903} and were adopted by several following papers \cite{DBLP:journals/corr/abs-2210-00720,DBLP:journals/corr/abs-2203-11171,DBLP:journals/corr/abs-2211-10435}.
We also used the P\textsc{a}L-style reasoning chains annotated by \citet{DBLP:journals/corr/abs-2211-10435}.

Prompting baselines without synthesis use all provided gold examples to construct prompts for inference.
S\textsc{ynthetic} \textsc{prompting} and its variants synthesize examples using the provided examples, and select 8 synthetic demonstrations based on in-cluster complexity, unless stated otherwise.

Seed examples and synthetic prompts are provided in the Supplementary Materials.

\subsection{Baselines}

\textbf{Direct Prompting}
Direct prompting \cite{DBLP:conf/nips/BrownMRSKDNSSAA20} prompts LLMs to directly generate answers with demonstrations of input-answer pairs.

\textbf{CoT Prompting}
Chain-of-thought prompting \cite{DBLP:journals/corr/abs-2201-11903} is effective in eliciting reasoning in LLMs, which prompts LLMs to generate natural language reasoning steps followed by an answer.

\textbf{P\textsc{a}L Prompting}
P\textsc{a}L prompting \cite{DBLP:journals/corr/abs-2211-10435}, a variant of chain-of-thought prompting, improves reasoning with structured code.
Figure \ref{fig:prompts_for_synthesis} (right) provides two examples.
It does not prompt LLMs to include final answers into completions; answers are obtained by executing the code.
This prompting technique has achieved state-of-the-art results on numerous reasoning tasks.

\textbf{Vanilla S\textsc{ynthetic} \textsc{prompting}}
This is a variant of S\textsc{ynthetic} \textsc{prompting}, which differs in that prompts used for question synthesis only consist of questions from seed examples.
In other words, new questions are synthesized by mimicking seed questions, without any other condition.

\begingroup
\setlength{\tabcolsep}{2.5pt} 
\renewcommand{\arraystretch}{1} 
\begin{table*}[htb]
    \centering
    \adjustbox{max width=.99\textwidth}{
    \begin{tabular}{
>{\columncolor[HTML]{FFFFFF}}c 
>{\columncolor[HTML]{DAE8FC}}c 
>{\columncolor[HTML]{DAE8FC}}c 
>{\columncolor[HTML]{DAE8FC}}c 
>{\columncolor[HTML]{FFFFFF}}c 
>{\columncolor[HTML]{FFFFFF}}c 
>{\columncolor[HTML]{FFFFFF}}c 
>{\columncolor[HTML]{DAE8FC}}c 
>{\columncolor[HTML]{DAE8FC}}c 
>{\columncolor[HTML]{DAE8FC}}c 
>{\columncolor[HTML]{FFFFFF}}c 
>{\columncolor[HTML]{FFFFFF}}c 
>{\columncolor[HTML]{FFFFFF}}c 
>{\columncolor[HTML]{DAE8FC}}c 
>{\columncolor[HTML]{DAE8FC}}c 
>{\columncolor[HTML]{DAE8FC}}c 
>{\columncolor[HTML]{FFFFFF}}c 
>{\columncolor[HTML]{FFFFFF}}c 
>{\columncolor[HTML]{DAE8FC}}c 
>{\columncolor[HTML]{DAE8FC}}c }
\hline
{\color[HTML]{000000} Datasets}                 & \multicolumn{3}{c}{\cellcolor[HTML]{DAE8FC}{\color[HTML]{000000} GSM8K}}                                                 & \multicolumn{3}{c}{\cellcolor[HTML]{FFFFFF}{\color[HTML]{000000} GSM-Hard}}                                                    & \multicolumn{3}{c}{\cellcolor[HTML]{DAE8FC}{\color[HTML]{000000} SVAMP}}                                                 & \multicolumn{3}{c}{\cellcolor[HTML]{FFFFFF}{\color[HTML]{000000} ASDiv}}                                        & \multicolumn{3}{c}{\cellcolor[HTML]{DAE8FC}{\color[HTML]{000000} SingleOp}}                                              & \multicolumn{2}{c}{\cellcolor[HTML]{FFFFFF}{\color[HTML]{000000} Colored Objects}} & \multicolumn{2}{c}{\cellcolor[HTML]{DAE8FC}{\color[HTML]{000000} Repeat Copy}}    \\ \hline
{\color[HTML]{000000} Previous Fine-tuned SOTA} & \multicolumn{3}{c}{\cellcolor[HTML]{DAE8FC}{\color[HTML]{000000} 55\textsuperscript{$\alpha$}}}                                                   & \multicolumn{3}{c}{\cellcolor[HTML]{FFFFFF}{\color[HTML]{000000} --}}                                                          & \multicolumn{3}{c}{\cellcolor[HTML]{DAE8FC}{\color[HTML]{000000} 57.4\textsuperscript{$\beta$}}}                                                 & \multicolumn{3}{c}{\cellcolor[HTML]{FFFFFF}{\color[HTML]{000000} 75.3\textsuperscript{$\gamma$}}}                                        & \multicolumn{3}{c}{\cellcolor[HTML]{DAE8FC}{\color[HTML]{000000} --}}                                                    & \multicolumn{2}{c}{\cellcolor[HTML]{FFFFFF}{\color[HTML]{000000} --}}              & \multicolumn{2}{c}{\cellcolor[HTML]{DAE8FC}{\color[HTML]{000000} --}}             \\
{\color[HTML]{000000} CoT\textsubscript{PaLM 540B}}            & \multicolumn{3}{c}{\cellcolor[HTML]{DAE8FC}{\color[HTML]{000000} 56.9}}                                                  & \multicolumn{3}{c}{\cellcolor[HTML]{FFFFFF}{\color[HTML]{000000} --}}                                                          & \multicolumn{3}{c}{\cellcolor[HTML]{DAE8FC}{\color[HTML]{000000} 79.0}}                                                  & \multicolumn{3}{c}{\cellcolor[HTML]{FFFFFF}{\color[HTML]{000000} 73.9}}                                         & \multicolumn{3}{c}{\cellcolor[HTML]{DAE8FC}{\color[HTML]{000000} --}}                                                    & \multicolumn{2}{c}{\cellcolor[HTML]{FFFFFF}{\color[HTML]{000000} --}}              & \multicolumn{2}{c}{\cellcolor[HTML]{DAE8FC}{\color[HTML]{000000} --}}             \\ \hline
{\color[HTML]{000000} \# Gold/Seed Examples}    & {\color[HTML]{000000} 2}             & {\color[HTML]{000000} 4}                   & {\color[HTML]{000000} 8}             & {\color[HTML]{000000} 2}             & {\color[HTML]{000000} 4}                   & {\color[HTML]{000000} 8}                   & {\color[HTML]{000000} 2}                   & {\color[HTML]{000000} 4}             & {\color[HTML]{000000} 8}             & {\color[HTML]{000000} 2}                   & {\color[HTML]{000000} 4}             & {\color[HTML]{000000} 8}    & {\color[HTML]{000000} 2}             & {\color[HTML]{000000} 4}             & {\color[HTML]{000000} 8}                   & {\color[HTML]{000000} 2}              & {\color[HTML]{000000} 4}                   & {\color[HTML]{000000} 2}                   & {\color[HTML]{000000} 4}             \\ \hline
\multicolumn{20}{c}{\cellcolor[HTML]{FFFFFF}{\color[HTML]{000000} \textit{Methods Using Gold Examples Only}}}                                                                                                                                                                                                                                                                                                                                                                                                                                                                                                                                                                                                                                                                                                                                                \\ \hline
{\color[HTML]{000000} Direct}                   & {\color[HTML]{000000} 15.4}          & {\color[HTML]{000000} 16.1}                & {\color[HTML]{000000} 16.8}          & {\color[HTML]{000000} 4.4}           & {\color[HTML]{000000} 4.3}                 & {\color[HTML]{000000} 4.4}                 & {\color[HTML]{000000} 67.0}                & {\color[HTML]{000000} 67.2}          & {\color[HTML]{000000} 68.3}          & {\color[HTML]{000000} 69.8}                & {\color[HTML]{000000} 69.8}          & {\color[HTML]{000000} 71.5} & {\color[HTML]{000000} 92.5}          & {\color[HTML]{000000} 92.2}          & {\color[HTML]{000000} 92.2}                & {\color[HTML]{000000} 58.7}           & {\color[HTML]{000000} 81.7}                & {\color[HTML]{000000} 31.3}                & {\color[HTML]{000000} 43.8}          \\
{\color[HTML]{000000} CoT}                      & {\color[HTML]{000000} 61.8}          & {\color[HTML]{000000} 62.2}                & {\color[HTML]{000000} 58.2}          & {\color[HTML]{000000} 23.6}          & {\color[HTML]{000000} 23.3}                & {\color[HTML]{000000} 23.0}                & {\color[HTML]{000000} 79.5}                & {\color[HTML]{000000} 78.2}          & {\color[HTML]{000000} 79.0}          & {\color[HTML]{000000} 75.1}                & {\color[HTML]{000000} 77.7}          & {\color[HTML]{000000} 80.2} & {\color[HTML]{000000} 90.7}          & {\color[HTML]{000000} 93.4}          & {\color[HTML]{000000} 93.4}                & {\color[HTML]{000000} 81.7}           & {\color[HTML]{000000} 88.8}                & {\color[HTML]{000000} 56.3}                & {\color[HTML]{000000} 62.5}          \\
{\color[HTML]{000000} P\textsc{a}L}                      & {\color[HTML]{000000} 72.5}          & {\color[HTML]{000000} 73.1}                & {\color[HTML]{000000} 71.8}          & {\color[HTML]{000000} 62.8}          & {\color[HTML]{000000} 62.9}                & {\color[HTML]{000000} 60.7}                & {\color[HTML]{000000} 80.2}                & {\color[HTML]{000000} 79.6}          & {\color[HTML]{000000} 81.8}          & {\color[HTML]{000000} 81.0}                & {\color[HTML]{000000} 79.4}          & {\color[HTML]{000000} 81.1} & {\color[HTML]{000000} 93.4}          & {\color[HTML]{000000} 92.5}          & {\color[HTML]{000000} 94.1}                & {\color[HTML]{000000} 82.7}           & {\color[HTML]{000000} 94.4}                & {\color[HTML]{000000} 71.9}                & {\color[HTML]{000000} 81.3}          \\ \hline
\multicolumn{20}{c}{\cellcolor[HTML]{FFFFFF}{\color[HTML]{000000} \textit{Methods Leveraging Synthetic Demonstrations}}}                                                                                                                                                                                                                                                                                                                                                                                                                                                                                                                                                                                                                                                                                                                                     \\ \hline
{\color[HTML]{000000} Vanilla S\textsc{ynthetic}}        & {\color[HTML]{000000} \textbf{72.6}} & {\color[HTML]{000000} 72.7}                & {\color[HTML]{000000} 71.6}          & {\color[HTML]{000000} 62.5}          & {\color[HTML]{000000} 59.3}                & {\color[HTML]{000000} 60.0}                & {\color[HTML]{000000} {\ul \textbf{83.5}}} & {\color[HTML]{000000} \textbf{80.6}} & {\color[HTML]{000000} 81.4}          & {\color[HTML]{000000} {\ul \textbf{81.5}}} & {\color[HTML]{000000} \textbf{80.8}} & {\color[HTML]{000000} 80.8} & {\color[HTML]{000000} 92.7}          & {\color[HTML]{000000} 91.5}          & {\color[HTML]{000000} \textbf{94.3}}       & {\color[HTML]{000000} \textbf{92.7}}  & {\color[HTML]{000000} 93.1}                & {\color[HTML]{000000} 53.1}                & {\color[HTML]{000000} 81.3}          \\
{\color[HTML]{000000} S\textsc{ynthetic}}                & {\color[HTML]{000000} \textbf{74.7}} & {\color[HTML]{000000} {\ul \textbf{75.3}}} & {\color[HTML]{000000} \textbf{73.9}} & {\color[HTML]{000000} \textbf{63.1}} & {\color[HTML]{000000} {\ul \textbf{64.7}}} & {\color[HTML]{000000} {\ul \textbf{64.7}}} & {\color[HTML]{000000} \textbf{82.0}}       & {\color[HTML]{000000} \textbf{80.5}} & {\color[HTML]{000000} \textbf{81.8}} & {\color[HTML]{000000} \textbf{81.1}}       & {\color[HTML]{000000} \textbf{80.6}} & {\color[HTML]{000000} 80.7} & {\color[HTML]{000000} \textbf{93.6}} & {\color[HTML]{000000} \textbf{93.4}} & {\color[HTML]{000000} {\ul \textbf{95.2}}} & {\color[HTML]{000000} \textbf{93.8}}  & {\color[HTML]{000000} {\ul \textbf{97.3}}} & {\color[HTML]{000000} {\ul \textbf{87.5}}} & {\color[HTML]{000000} \textbf{84.4}} \\ \hline
    \end{tabular}
    }
    \caption{
    Accuracies on numerical reasoning, symbolic reasoning, and algorithmic reasoning datasets.
    The previous fine-tuned state-of-the-art models are: $\alpha$: \cite{DBLP:journals/corr/abs-2110-14168}; $\beta$: \cite{DBLP:journals/corr/abs-2201-11473}; $\gamma$: \cite{DBLP:conf/acl/MiaoLS20}.
    Accuracies of CoT prompting with PaLM 540B are from \citet{DBLP:journals/corr/abs-2201-11903}.
    P\textsc{a}L prompting is the previous state-of-the-art method, which uses provided gold examples as demonstrations during inference.
    Both S\textsc{ynthetic prompting} and its vanilla version use gold examples as seeds to automatically synthesize more examples by prompting an LLM; during inference, demonstrations are selected from self-synthesized examples to better elicit reasoning in the LLM itself.
    }
    \label{tab:main}
\end{table*}
\endgroup

\subsection{Implementation Details}
We adopted P\textsc{a}L-style reasoning chains which are structured code with comments being natural language reasoning step.
\texttt{text-davinci-003} version of InstructGPT \cite{DBLP:journals/corr/abs-2203-02155} was used as our backend LLM for both synthesis and inference.
We used top-p sampling \cite{DBLP:conf/iclr/HoltzmanBDFC20} for synthesis with temperature set to 0.7, and used greedy decoding for inference with temperature set to 0.
All numerical reasoning datasets share one set of seed examples either randomly sampled from GSM8K (when the number of seeds is 2 or 4) or from \citet{DBLP:journals/corr/abs-2201-11903} (when the number of seeds is 8).
For datasets of the other tasks, seeds were randomly sampled from their own datasets.
We annotated seed examples with both CoT-style reasoning chains and P\textsc{a}L-style reasoning chains manually, following their annotation protocols.
Annotations are provided in the Supplementary Materials.
For each set of seed examples, we synthesized more examples by repeating backward-forward synthesis for 1,000 times.
Target complexities range from the lowest complexity of seed examples to the highest one plus $c$; $c$ was set to 4 for numerical reasoning and 2 on the other datasets.
In forward synthesis, the number of reasoning chains sampled for each question was 3.
The encoder used for clustering was \texttt{all-mpnet-base-v2}.

\subsection{Main Results}
As shown by Table \ref{tab:main}, S\textsc{ynthetic prompting} consistently outperforms P\textsc{a}L prompting by up to absolute 15.6\%, indicating that self-synthesized demonstrations can be leveraged to better elicit reasoning in the LLM itself, surpassing the performance of using seed demonstrations only.

Though vanilla S\textsc{ynthetic prompting} also uses synthetic demonstrations, it fails to consistently improve over P\textsc{a}L prompting.
On GSM8K and GSM-Hard which contain questions requiring complex deductions, vanilla S\textsc{ynthetic prompting} can barely improve over P\textsc{a}L prompting, as it does not explicitly control the reasoning complexities of synthetic examples and tends to synthesize examples that are similar to seed examples in terms of complexities and informativeness. 
Notably, vanilla S\textsc{ynthetic prompting} significantly underperforms P\textsc{a}L prompting on Repeat Copy with 2 seed examples.
We found that 2 selected demonstrations have ill-formed questions, e.g., \textit{Repeat the sentence "The sun is bright" five times, with a different emphasis on a different word each time}.
This may be because questions are synthesized without explicit awareness of their reasoning chains.
Section \ref{sec:qg} shows the benefits of controlling question synthesis with various conditions.

We also observe that increasing the number of seed examples from 2 to 8 does not significantly improve performance, especially on GSM8K and Repeat Copy.
Two possible reasons are as follows:
(1) Example synthesis are biased by seed examples. With limited seeds, it is possible that synthesized examples are not diverse enough, and are still helpless on some portion of test questions.
(2) Though our proposed demonstration selection scheme is effective (see analysis in Section \ref{sec:demo_selection}), it is probably suboptimal, failing to make the best of synthesized examples.

\subsection{Ablation Studies}
We mainly conducted ablation studies on GSM8K and the Colored Objects task.

\subsubsection{Conditions Used for Question Synthesis}
\label{sec:qg}
\begingroup
\setlength{\tabcolsep}{3pt} 
\renewcommand{\arraystretch}{1} 
\begin{table}[htb]
    \centering
    \adjustbox{max width=.47\textwidth}{
    \begin{tabular}{lcccc}
        \hline
        Datasets & \multicolumn{2}{c}{GSM8K} & \multicolumn{2}{c}{Colored Objects} \\
        \cmidrule(lr){2-3} \cmidrule(lr){4-5}
        \# Seed Examples & 2 & 4 & 2 & 4 \\
        \hline
        S\textsc{ynthetic prompting} & \textbf{74.7} & \textbf{75.3} & \textbf{93.8} & \textbf{97.3} \\
        w/o Topic Word & 73.9 & 73.2 & 80.2 & 92.9 \\
        w/o Target Complexity & 71.6 & 72.0 & 93.1 & 94.5  \\
        w/o Reasoning Chain & 72.9 & 73.8 & 86.0 & 92.2 \\
	\hline
    \end{tabular}
    }
    \caption{Analysis of how different conditions used in backward question synthesis affect inference performance.}
    \label{tab:qg_ablation}
\end{table}
\endgroup

\begingroup
\setlength{\tabcolsep}{3pt} 
\renewcommand{\arraystretch}{1} 
\begin{table}[htb]
    \centering
    \adjustbox{max width=0.49\textwidth}{
    \begin{tabular}{lccc}
        \hline
        Method & Diversity & Complexity & Correctness \\
        \hline
        S\textsc{ynthetic prompting} & 0.68 & 8.23 / 12.5 & 100.0\% \\
        w/o Topic Word & 0.85 & 9.4 / 13.4 & 100.0\% \\
        w/o Target Complexity & 0.73 & 5.0 / 7.6 & 87.5\%  \\
        w/o Reasoning Chain & 0.68 & 5.6 / 12.3 & 37.5\% \\
	\hline
    \end{tabular}
    }
    \caption{
    Analysis on GSM8K with 4 seed examples, about how conditions used in backward question synthesis affect quality of synthesized examples.
    \texttt{Diversity} is measured by pair-wise cosine similarity among synthetic examples.
    \texttt{Complexity} measures the average reasoning complexities of synthetic examples (left) and selected demonstrations (right), separated by \texttt{/}.
    \texttt{Correctness} is the portion of selected synthetic demonstrations that are correct.
    }
    \label{tab:qg_ablation_analysis}
\end{table}
\endgroup

In backward synthesis, we ask the LLM to sample a question conditioned on a topic, a target complexity, and a sampled reasoning chain.
To analyze the effect of each condition on question synthesis, we removed corresponding lines in the prompts.
Notably, when removing the target complexity, number markers of reasoning steps are also removed.
As shown by Table \ref{tab:qg_ablation}, removing any condition leads to degraded model performance on both GSM8K and Colored Objects.

We further investigated how different conditions affect the quality of synthetic examples, in terms of (1) \textbf{diversity}, measured by the maximum pair-wise cosine similarity between a synthetic example and the others on average, (2) \textbf{complexity}, measured by the average number of reasoning steps, (3) and \textbf{correctness}, measured by the portion of demonstrations used for inference that are correct.
Table \ref{tab:qg_ablation_analysis} presents the analysis on GSM8K.
\textbf{Removing topic words} results in less diverse synthetic examples. Reasoning patterns of selected demonstrations are limited too; although all demonstrations are correct, 62.5\% of questions revolve around \textit{discount} or \textit{tax}.
\textbf{Removing target complexities} produces much simpler synthetic examples.
Synthesizing questions \textbf{without conditioned on reasoning chains} affects correctness negatively; 62.5\% are flawed, 80\% of which are unanswerable, e.g.,

\textit{An image is represented by a 4x4 matrix of integers. Each cell of the matrix contains a single integer between 0 and 255. What is the average value of all the integers in the matrix?}.

Notably, though we also include target complexities into the prompts when synthesizing questions without conditioned on reasoning chains, the resulting questions tend to require less reasoning steps than S\textsc{ynthetic prompting}, indicating that conditioning on numbered reasoning steps can control reasoning complexities better.

\subsubsection{Schemes of Demonstration Selection}
\label{sec:demo_selection}
\begingroup
\setlength{\tabcolsep}{3pt} 
\renewcommand{\arraystretch}{1} 
\begin{table}[htb]
    \centering
    \adjustbox{max width=.47\textwidth}{
    \begin{tabular}{lcccc}
        \hline
        Datasets & \multicolumn{2}{c}{GSM8K} & \multicolumn{2}{c}{Colored Objects} \\
        \cmidrule(lr){2-3} \cmidrule(lr){4-5}
        \# Seed Examples & 2 & 4 & 2 & 4 \\
        \hline
        Random & 73.4 & 73.1 & 83.0 & 90.0  \\
        Cluster Centroid & 73.1 & 73.0 & 91.2 & 93.6 \\
        Similarity & 72.9 & 74.0 & 87.0 & 94.2 \\
        In-Cluster Similarity & 72.9 & 73.2 & 89.3 & 95.4 \\
        Complexity & 74.1 & 74.3 & 92.7 & \textbf{97.5} \\
        In-Cluster Complexity & \textbf{74.7} & \textbf{75.3} & \textbf{93.8} & 97.3 \\
	\hline
    \end{tabular}
    }
    \caption{Accuracies with different schemes of demonstration selection.}
    \label{tab:demo_selection}
\end{table}
\endgroup
To make good use of synthesized examples, having an effecitve selection scheme matters.
We evaluated the following 6 selection schemes.
(1) \textbf{Random}: randomly selects demonstrations;
(2) \textbf{Cluster Centroid}: selects the example closest to each cluster centroid;
(3) \textbf{Similarity}: retrieves the most similar examples according to cosine similarity;
(4) \textbf{In-Cluster Similarity}: select the most similar example from each cluster;
(5) \textbf{Complexity}: selects the examples with the most reasoning steps;
(6) \textbf{In-Cluster Complexity}: selects the most complex example from each cluster.

Table \ref{tab:demo_selection} presents the comparisons.
Though most selection schemes achieve better performance than P\textsc{a}L prompting, complexity-based selection schemes are the most effective on the two reasoning tasks, with some other schemes like Random lagging far behind.
Our proposed In-Cluster Complexity outperforms Complexity, showing the benefits of using diverse and complex demonstrations.

\begingroup
\setlength{\tabcolsep}{3pt} 
\renewcommand{\arraystretch}{1} 
\begin{table*}[htb]
    \centering
    \adjustbox{max width=0.99\textwidth}{
    \begin{tabular}{p{0.99\textwidth}}
        \hline
        \multicolumn{1}{c}{\textbf{Seed Examples}} \\
        \hline
        \textbf{Question:} Carol has 20 signatures in her book, and Jennifer has 44. The sisters have three more weeks of summer vacation, and they decide they want to reach 100 signatures between them by the end of the summer. How many signatures do the sisters need to collect to reach their goal? \\
        \\
        \textbf{Question:} A team of 4 painters worked on a mansion for 3/8ths of a day every day for 3 weeks. How many hours of work did each painter put in? \\
        \hline
        \multicolumn{1}{c}{\textbf{Synthetic Demonstrations from Vanilla S\textsc{ynthetic prompting}}} \\
        \hline
        \textbf{Question:} 80 people attended a party. 40 of them were men and 40 were women. Of the men, 10 were married and of the women, 20 were married. How many single people were at the party? \\
        \\
        \textbf{Question:} The team of 4 painters worked on the mansion for 3/8ths of a day every day for 3 weeks. If each painter was paid \$25 per hour, how much money did the team of painters earn in total? \\
        \\
        \textbf{Question (Unanswerable):} 80 students took a quiz in Mrs. Smith's math class. The average score was 75\%. How many students scored a 95\% or higher? \\
	\hline
        \multicolumn{1}{c}{\textbf{Synthetic Demonstrations from S\textsc{ynthetic prompting}}} \\
        \hline
        \textbf{Given Topic Word: } idea \\
        \textbf{Question:} If 5 people each have 10 ideas, with 5 of those ideas being innovative and taking 2 minutes each to develop, and the other 5 ideas being non-innovative and taking 1 minute each to develop, how many minutes will it take all 5 people to develop all 10 ideas? \\
        \\
        \textbf{Given Topic Word: } office \\
        \textbf{Question:} The office is 20 feet wide, 30 feet long and 10 feet high. It has two windows that are each 5 feet wide and 6 feet high, and one door that is 3 feet wide and 8 feet high. What is the total area of the office walls? \\
        \\
        \textbf{Given Topic Word: } gallery \\
        \textbf{Question:} A gallery has 10 paintings, 9 sculptures, 6 photos, and 4 mixed media pieces. The painting is \$200, the sculpture is \$500, the photo is \$100 and the mixed media piece is \$150. You get a 15\% discount and you have to pay 5\% tax. How much will you pay in total? \\
        \hline
    \end{tabular}
    }
    \caption{
    A set of seed examples from GSM8K, as well as synthetic demonstrations selected by S\textsc{ynthetic prompting} and its vanilla version, respectively.
    We only show questions for brevity, as their reasoning chains are correct, except that the third question from vanilla S\textsc{ynthetic prompting} is unanswerable.
    }
    \label{tab:synthetic_samples}
\end{table*}
\endgroup

\subsubsection{Sensitivity to Seed Examples}
\begin{figure}[htb]
    \centering
    \includegraphics[width=0.49\textwidth]{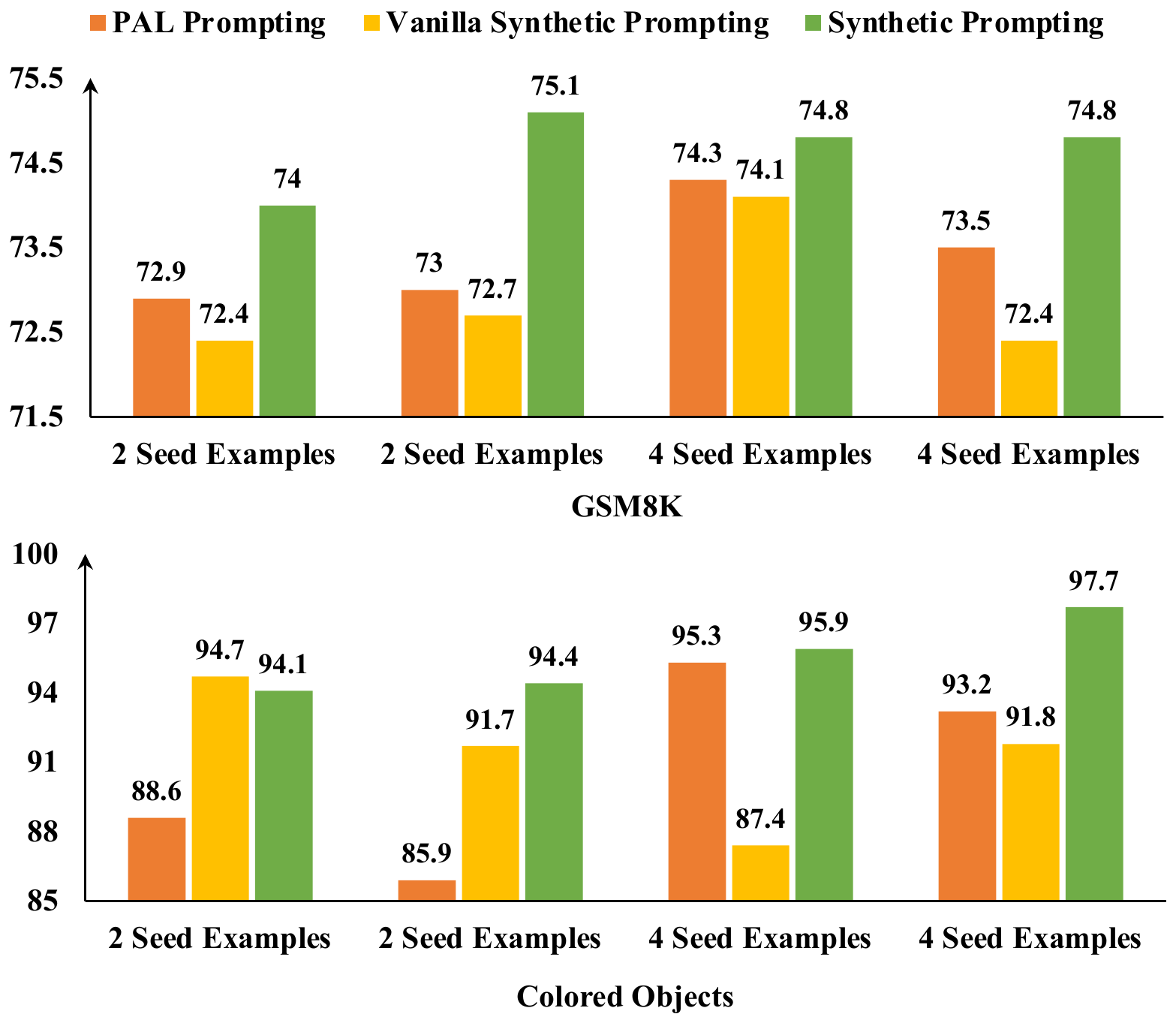}
    \caption{
    Sensitivity analysis on GSM8K and Colored Objects.
    We experimented with another two random sets of seed examples of size 2 and 4 for each dataset.
    }
    \label{fig:sensitivity}
\end{figure}
To investigate how sensitive S\textsc{ynthetic prompting} is to seed examples, we repeated experiments on another two random sets of seed examples.
Figure \ref{fig:sensitivity} demonstrates our sensitivity analysis.
S\textsc{ynthetic prompting} consistently outperforms P\textsc{a}L prompting on different runs.
However, we observed that seed examples with better P\textsc{a}L prompting performance does not necessarily lead to better S\textsc{ynthetic prompting} performance. 

\subsection{Comparison with Selecting from Training Examples}
\label{sec:gap}
To measure the performance gap between using synthetic demonstrations and using gold demonstrations from a large set of carefully-curated examples, we selected 8 demonstrations from the training set of GSM8K with the two complexity-based selection schemes (i.e., Complexity and In-Cluster Complexity in Section \ref{sec:demo_selection}), respectively.
As the training examples were annotated with natural language reasoning chains (CoT-style reasoning), we measured the numbers of natural language reasoning steps as reasoning complexities for complexity-based selection, and manually annotated selected examples with P\textsc{a}L-style reasoning chains for P\textsc{a}L prompting.
As the training examples of GSM8K are diverse, both Complexity and In-Cluster Complexity select diverse and informative demonstrations, and yield an accuracy of 77.0\% on the test set of GSM8K, surpassing our accuracy of 75.3\% by absolute 1.7\%.
As shown in the Supplementary Materials, compared with our synthetic demonstrations,
the selected gold demonstrations are more logically complex with less straightforward reasoning, which may be more informative to LLMs.

Notably, using the 8 simple demonstrations from \citet{DBLP:journals/corr/abs-2201-11903} that were manually crafted without prompt engineering results in an even lower accuracy of 71.8\%.
This indicates that demonstrations indeed matters.
Under scenarios with access to only limited and possibly-simple examples, automatically synthesizing examples for selecting more effective demonstrations serves as a promising way to elicit better reasoning in LLMs.

\subsection{Quality Analysis of Synthetic Examples}
To investigate the quality of synthesized examples, we conducted manual evaluation on GSM8K.
We evaluated 25 random examples synthesized by S\textsc{ynthetic prompting} and vanilla S\textsc{ynthetic prompting}, respectively.
Compared with vanilla S\textsc{ynthetic prompting}, S\textsc{ynthetic prompting} synthesizes questions of higher complexities (8.3 vs. 5.4) and also with lower error rate (8\% vs. 24\%).

We further analyze the quality of selected synthetic demonstrations.
For S\textsc{ynthetic prompting}, all selected demonstrations are correct, while its vanilla version has one unanswerable question and another one with wrong reasoning.

Table \ref{tab:synthetic_samples} shows two seed examples, as well as some synthetic demonstrations.
For vanilla S\textsc{ynthetic prompting}, the first two questions are logically close to seed questions, and the third one is unanswerable.
With S\textsc{ynthetic prompting}, the LLM can synthesize on-topic questions requiring novel reasoning patterns, e.g., the second question about \texttt{office} requires geometric reasoning.

\section{Conclusion}

We introduce S\textsc{ynthetic prompting}, a novel technique for reasoning with large language models using few examples, that differs from previous work by using the models as generators of additional examples besides as consumers of in-context demonstrations. We show that by prompting a large language model to synthesize more examples, we can improve its reasoning performance on numerical, symbolic, and algorithmic tasks, compared to existing prompting methods such as chain-of-thought prompting and P\textsc{a}L prompting.



\bibliography{cot,datasets,in_context_learning,instruction_tuning,large_language_models,others}
\bibliographystyle{icml2023}

\newpage
\appendix
\onecolumn



\end{document}